\documentclass[letterpaper]{article} 
\usepackage{aaai2027}  
\usepackage[hyphens]{url}  
\usepackage{graphicx} 
\usepackage{natbib}  
\usepackage{caption} 

\usepackage{xcolor}
\usepackage{algorithm}
\usepackage{algorithmic}
\usepackage{amsfonts}
\usepackage{amsmath}
\usepackage{amsthm}

\usepackage{newfloat}
\usepackage{listings}

\usepackage{xcolor}

\DeclareCaptionStyle{ruled}{labelfont=normalfont,labelsep=colon,strut=off} 
\floatstyle{ruled}
\newfloat{listing}{tb}{lst}{}
\floatname{listing}{Listing}

\usepackage{booktabs}

\title{Can Attack Difficulty Be Characterized Before Optimization? \\
A Study of Pre-optimization Difficulty in Person-Vanishing Attacks}
\author{
    Jingyao Xu\textsuperscript{\rm 1}\equalcontrib,
    Dongdong Wang\textsuperscript{\rm 2}\equalcontrib,
    Siyang Lu\textsuperscript{\rm 1}\corresponding
}
\affiliations{
    \textsuperscript{\rm 1}School of Computer Science and Technology, Beijing Jiaotong University\\
    \textsuperscript{\rm 2}College of Design, Construction, and Planning, University of Florida\\

    \{jingyaoxu, sylu\}@bjtu.edu.cn, dongdongwang@ufl.edu
}

\makeatletter
\renewcommand{\copyright@on}{}
\makeatother

\begin{document}

\maketitle

\begin{abstract}

Adversarial attacks against object detectors are traditionally studied from an optimization perspective, where attack difficulty is regarded as an outcome observed only after adversarial optimization. This raises a fundamental question: \emph{can the relative attack difficulty of different inputs be characterized before optimization begins?} In this paper, we investigate this question for person-vanishing attacks by introducing the concept of pre-optimization attack difficulty, which captures intrinsic differences in optimization effort across input images. To estimate this latent difficulty before optimization, we propose Quad-CLEVER, an efficient geometry-based estimator derived from a quadratic approximation of the local person-vanishing margin along the most attack-relevant direction. Extensive experiments across multiple attack algorithms demonstrate that Quad-CLEVER consistently correlates with the observed optimization cost, providing empirical evidence that attack difficulty exhibits a predictable pre-optimization structure. Building upon this finding, we further propose a difficulty-aware attack framework that leverages the estimated difficulty to adaptively allocate optimization budgets for a base attack under a fixed computational budget. On BDD100K, the proposed framework improves the image-level attack success rate by up to 5.78$\%$ while reducing the average optimization cost by up to 11.42 iterations. On the more challenging EventPed dataset, it saves 2.25 optimization iterations while maintaining comparable attack performance. These results demonstrate that attack difficulty can be meaningfully estimated before optimization and that exploiting such estimates enables more computationally efficient adversarial attacks.
\end{abstract}

\section{Introduction}

Recent advances in artificial intelligence have accelerated the deployment of autonomous driving systems, where object detection serves as a fundamental perception component for recognizing surrounding traffic participants and supporting downstream driving decisions. Among all detection targets, pedestrians are particularly safety-critical because failing to detect even a single pedestrian may lead to catastrophic consequences~\cite{el2020pedestrian,lyssenko2024safety}. Consequently, understanding the vulnerability of pedestrian detection is essential for improving the safety and reliability of autonomous driving systems~\cite{zhang2021evaluating}.

Person-vanishing attacks deliberately suppress pedestrian detections and have become an important benchmark for evaluating the robustness of object detectors~\cite{wu2020making,hu2022adversarial,guesmi2024dap,lin2025adversarial}. Existing studies have primarily focused on developing stronger attack objectives, more effective optimization algorithms, and higher attack success rates. While these efforts have substantially advanced attack capability, they largely view attack difficulty as an outcome of the optimization process itself. As a result, comparatively little attention has been devoted to understanding whether attack difficulty is merely produced during optimization or reflects a property that can be characterized prior to optimization.

This observation motivates a fundamental scientific question:

\begin{quote}
\emph{Does person-vanishing attack difficulty emerge solely from adversarial optimization, or does it contain an observable input-dependent component that exists before optimization begins?}
\end{quote}

Answering this question has important implications. If attack difficulty is determined entirely by the optimization procedure, the same image may exhibit arbitrarily different difficulty under different attack algorithms, making meaningful pre-optimization estimation impossible. Conversely, if attack difficulty contains an observable input-dependent component, two testable consequences should follow. First, the relative difficulty of images should be at least partially preserved across different attack algorithms. Second, attack difficulty should be estimable directly from clean inputs before any adversarial optimization is performed. Establishing these properties would shift the study of person-vanishing attacks from exclusively designing stronger optimization procedures toward understanding the structure of attack difficulty itself.

Motivated by this question, we investigate whether data difficulty in person-vanishing attacks can be estimated before adversarial optimization. Rather than treating attack difficulty solely as a retrospective quantity observed after optimization, we investigate whether the relative attack difficulty across data samples can be characterized directly from clean inputs. To this end, we introduce Quad-CLEVER, a pre-optimization estimator that predicts the relative attack difficulty of clean inputs without executing adversarial optimization. We further evaluate whether the estimated difficulty is consistent with observed attack difficulty across different attack algorithms and whether it supports more effective attack decision making through adaptive budget routing.

Rather than asking how to construct stronger person-vanishing attacks, this work asks whether attack difficulty itself can be systematically characterized before adversarial optimization. Our contributions are summarized as follows.

\begin{itemize}

\item We investigate whether attack difficulty in person-vanishing attacks can be characterized before adversarial optimization. Through extensive experiments, we provide empirical evidence that the relative attack difficulty across input samples exhibits a predictable pre-optimization structure.

\item We propose Quad-CLEVER, an efficient geometry-based estimator that predicts the relative attack difficulty of clean inputs before adversarial optimization through a quadratic approximation of the local person-vanishing margin.

\item We develop a difficulty-aware attack framework that leverages the estimated difficulty to adaptively allocate optimization budgets, enabling more efficient attacks under a fixed computational budget.

\item We demonstrate that Quad-CLEVER generalizes across multiple attack algorithms and is consistently correlated with the observed optimization cost. The resulting framework achieves higher attack efficiency while maintaining competitive attack performance on two challenging benchmarks.

\end{itemize}

    \begin{figure}[t]
        \centering
        \includegraphics[width=1\columnwidth]{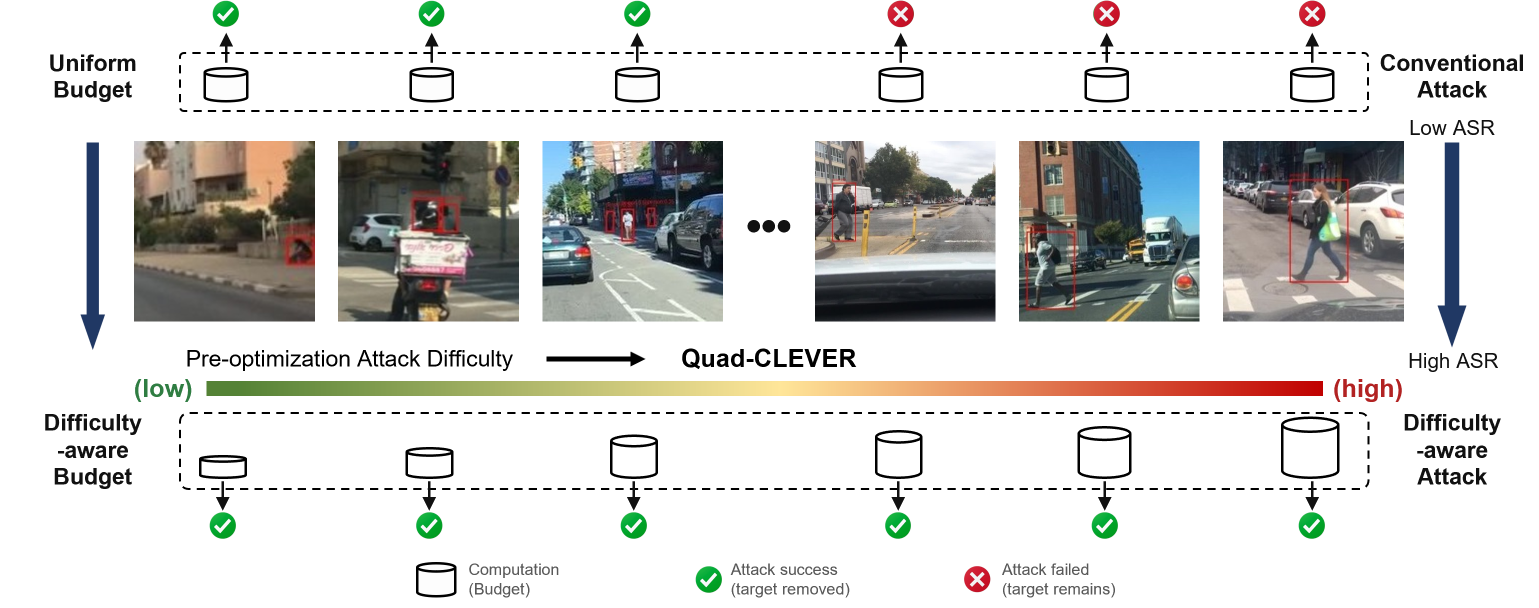}
        \caption{
        Images exhibit substantially different attack difficulty, making a uniform attack budget inefficient: easy images may receive redundant computation, while hard images remain under-optimized
        }
        \label{fig:overview}
        \vspace{-5pt}
    \end{figure}

\section{Related Work}

\subsection{Person-vanishing attacks on object detectors}

Adversarial attacks on object detectors \cite{madry2017towards,xie2017adversarial,chow2020adversarial} are more complex than classification attacks because detectors jointly predict object locations, categories, and confidence scores. Early physical studies showed that adversarial patches could suppress person detections under realistic viewing conditions \cite{thys2019fooling}. Subsequent work explored different vulnerabilities within the detection pipeline. Daedalus attacks the non-maximum suppression stage by manipulating bounding-box predictions \cite{wang2021daedalus}. DetectSec further provides a systematic framework for evaluating detector robustness under different adversarial attacks and model architectures \cite{du2022detects}. More recently, AFOG uses learnable attention to identify vulnerable image regions and attacks both transformer- and CNN-based object detectors \cite{yahn2025adversarial}. While these efforts have substantially advanced attack capability, attack difficulty is typically assessed retrospectively through optimization-dependent outcomes, such as attack success rate, residual detection confidence, perturbation magnitude, or the number of optimization iterations or queries. Whether relative person-vanishing difficulty can be characterized directly from clean inputs before optimization has received comparatively limited attention.

\subsection{Sample-level adversarial vulnerability}
Adversarial vulnerability can vary substantially across input samples. Source-image selection has been shown to influence attack success, transferability, and the perturbation required to generate adversarial examples \cite{ozbulak2021selection}. Raina and Gales further studied sample attackability in image classification by predicting whether individual samples can be attacked within a given perturbation range \cite{raina2023identifying}. Recent work has also examined sample difficulty in adversarial training. Liu et al.~\cite{liu2024impact} introduced an instance-level difficulty metric and showed that hard adversarial instances contribute disproportionately to robust overfitting. These findings provide evidence that adversarial behavior contains meaningful sample-dependent structure rather than being determined entirely by an attack algorithm or a uniform perturbation budget. Our work differs from these studies. We study person-vanishing attacks on structured object detectors rather than label-changing attacks on image classifiers; Our target quantity is the optimization effort required by an attack, rather than transferability or minimum perturbation magnitude; Quad-CLEVER is derived analytically from the clean detector response and does not require an auxiliary predictor trained with attack-generated difficulty labels. It is therefore designed to expose pre-optimization difficulty directly from the local geometry of each clean input.

\subsection{Robustness estimation}
A related research direction characterizes adversarial vulnerability through the local geometry of prediction functions. DeepFool estimates adversarial perturbations by repeatedly linearizing a classifier and approaching its decision boundary \cite{moosavi2016deepfool}. Analyses of decision-boundary geometry further show that both boundary distance and curvature influence adversarial robustness \cite{fawzi2016robustness}. CLEVER introduced an attack-independent robustness score based on local Lipschitz estimation and extreme value theory \cite{weng2018evaluating}. Its second-order extension incorporates local curvature for twice-differentiable prediction functions \cite{weng2018extensions}. These studies demonstrate that first- and second-order information around clean inputs can provide useful estimates of adversarial vulnerability. Quad-CLEVER transfers the local-geometric perspective to this structured setting but addresses a different prediction target. The resulting score is not presented as a certified robustness radius. Instead, it serves as a pre-optimization proxy for the relative computational effort subsequently observed across multiple attack algorithms.

\subsection{Adaptive adversarial optimization}

Sample-specific adversarial properties have also been used to adapt adversarial training. MMA training estimates individual sample margins and applies different perturbation radii according to the shortest successful adversarial perturbation \cite{ding2018mma}. Friendly Adversarial Training terminates inner-loop optimization after a suitable misclassified example has been found, reducing unnecessary attack iterations \cite{zhang2020attacks}. Geometry-Aware Instance-Reweighted Adversarial Training uses the number of attack steps required to misclassify a sample as an indicator for assigning training weights \cite{zhang2020geometry}. Customized Adversarial Training similarly assigns sample-dependent perturbation strengths rather than applying a uniform radius to all training inputs \cite{cheng2020cat}.

\section{Preliminaries}

\subsection{Attack Difficulty}

Let $f$ denote an object detector, $x\in\mathcal X$ an input image, and
$A\in\mathcal A$ an adversarial attack algorithm. We denote by
$C(x,A;f)$ the optimization cost required by attack algorithm $A$ to
successfully suppress all pedestrian detections in $x$. Depending on the
attack algorithm, the optimization cost may correspond to the number of
optimization iterations, optimization time, query count, or other
measures of optimization effort. Since the observed optimization cost varies across attack algorithms, we
define the attack difficulty of an input as the expected optimization
cost over an attack family,
\begin{equation}
\mathcal{K}(x;f)=\mathbb{E}_{A\sim\mathcal P(\mathcal A)}[C(x,A;f)],
\label{eq:attack_difficulty}
\end{equation}
where $\mathcal P(\mathcal A)$ denotes a distribution over attack
algorithms. A larger value of $K(x;f)$ indicates that the input requires,
on average, greater optimization effort to be successfully attacked.

\subsection{Bayesian Perspective of Attack Difficulty}

The intrinsic attack difficulty of an input is not directly observable before adversarial optimization. We therefore model it as a latent variable $D(x)$ and denote the observable pre-optimization information derived from the clean input and target model by $E(x)$. From a Bayesian perspective, the latent difficulty is inferred from the posterior
\begin{equation}
P(D(x)\mid E(x))
=
\frac{P(E(x)\mid D(x))P(D(x))}{P(E(x))}.
\end{equation}
This formulation suggests that pre-optimization difficulty estimation should rely on observable evidence rather than directly measuring the latent difficulty. Inspired by the local robustness perspective of CLEVER, we hypothesize that the local decision geometry surrounding the clean input provides informative evidence of the subsequent optimization difficulty. This motivates the search for a computationally efficient geometry-based evidence that can be extracted before adversarial optimization.

\subsection{CLEVER}

The Cross Lipschitz Extreme Value for nEtwork Robustness (CLEVER)~\cite{weng2018evaluating,weng2018extensions} score is an attack-independent robustness metric for neural networks. Unlike attack-based evaluation, CLEVER estimates a lower bound on the minimum adversarial perturbation using only the local geometry of the decision boundary, without performing adversarial optimization. Owing to its strong theoretical foundation and extensive empirical validation, CLEVER has become a reliable and well-established measure of local adversarial robustness. In this work, we do not use CLEVER to measure robustness. Instead, we build upon its pre-optimization geometric formulation and investigate whether similar geometry-derived quantities can serve as observable evidence for estimating latent attack difficulty before adversarial optimization.
\section{Methodology}

Following the Bayesian formulation in Section~3, we instantiate the geometry-based evidence using Quad-CLEVER and develop a difficulty-aware attack framework for efficient person-vanishing attacks. The framework consists of two stages. First, we instantiate the geometry-based evidence using Quad-CLEVER, a quadratic approximation of the detector's local margin geometry, and infer the latent attack difficulty from the resulting pre-optimization evidence. Second, the inferred difficulty is used to adaptively allocate the optimization budget of a base attack, enabling more efficient attacks under a fixed computational budget.


\subsection{Quad-CLEVER for Difficulty Estimation}

Quad-CLEVER estimates the relative optimization difficulty of
suppressing each detected person before adversarial optimization.
Inspired by the local geometric formulation of
CLEVER, we replace its sampling-based local
Lipschitz estimation with a second-order approximation of the detector
margin, yielding a closed-form estimate of the local
margin-crossing distance.

Let $\tilde{s}_o(x)$ denote the differentiable confidence score of the
pre-NMS hypothesis matched to target person $o$, and let $\tau$ denote
the detector threshold. Following CLEVER, we define the
person-vanishing margin as

\begin{equation}
g_o(x)
=
\log \tilde{s}_o(x)-\log\tau,
\label{eq:margin}
\end{equation}
where $g_o(x)>0$ indicates that the matched hypothesis remains
detectable.

Instead of estimating the local Lipschitz constant through repeated
gradient sampling, we directly model the local margin geometry. For each
target object, let

\begin{equation}
g_{0,o}=g_o(x),\qquad
b_o=\|\nabla_x g_o(x)\|_2,\qquad
u_o=-\frac{\nabla_x g_o(x)}{b_o},
\end{equation}
where $b_o$ is the gradient magnitude and $u_o$ is the direction of the
steepest margin decrease. The corresponding directional curvature is

\begin{equation}
\kappa_o
=
u_o^\top
\nabla_x^2g_o(x)
u_o.
\label{eq:kappa}
\end{equation}

Using a second-order Taylor approximation along $u_o$, the local margin
is approximated as

\begin{equation}
g_o(x+r u_o)
\approx
g_{0,o}
-
b_o r
+
\frac12\kappa_o r^2,
\qquad r\ge0.
\label{eq:taylor}
\end{equation}

The local margin-crossing distance is obtained by solving the quadratic
approximation in Eq.~(\ref{eq:taylor}), resulting in the proposed
Quad-CLEVER estimator

\begin{equation}
\mathrm{Quad\mbox{-}CLEVER}(x,o)
=
\frac{2g_{0,o}}
{b_o+\sqrt{b_o^2-2\kappa_o g_{0,o}}},
\label{eq:quadclever}
\end{equation}
which is algebraically equivalent to
$(b_o-\sqrt{b_o^2-2\kappa_o g_{0,o}})/\kappa_o$
while remaining numerically stable as
$\kappa_o\rightarrow0$, where it naturally reduces to the first-order
estimate $g_{0,o}/b_o$.

A smaller Quad-CLEVER score indicates that the detector margin reaches
the decision boundary under a smaller local perturbation and is
therefore expected to require less optimization effort. Compared with
sampling-based CLEVER, Quad-CLEVER requires only one gradient evaluation
and one Hessian-vector product, eliminating the sampling overhead of
local Lipschitz estimation.

Since an image may contain multiple target persons, we aggregate the
object-level estimates into an image-level difficulty score,

\begin{equation}
\widehat D(x;f)
=
\min_{o\in\mathcal O(x)}
\mathrm{Quad\mbox{-}CLEVER}(x,o),
\label{eq:difficulty}
\end{equation}
where $\mathcal O(x)$ denotes the detected target persons. Since a
person-vanishing attack succeeds once any target disappears, the minimum
naturally characterizes the difficulty of the attack objective.

\subsection{Difficulty-Aware Attack}

Having established Quad-CLEVER as an effective pre-optimization estimator of attack difficulty, we next investigate whether the estimated difficulty can be exploited to improve attack efficiency. To this end, we propose a difficulty-aware attack framework that consists of three steps: (1) estimating the image-level attack difficulty using Quad-CLEVER; (2) allocating an image-level optimization budget through a budget allocation policy; and (3) executing a base attack with the assigned budget. Algorithm~\ref{alg:framework} summarizes the overall framework.

\begin{algorithm}[h]
\caption{Difficulty-Aware Attack Framework}
\label{alg:framework}
\begin{algorithmic}[1]
\REQUIRE Inputs $\mathcal X$, detector $f$, base attack $\mathcal A$, total optimization budget $B$
\FOR{each input $x\in\mathcal X$}
    \STATE Estimate attack difficulty:
    $\hat D(x)\leftarrow\Phi(x,f)$
    \STATE Allocate optimization budget:
    $K(x)\leftarrow\pi(\hat D(x),B)$
    \STATE Generate adversarial example:
    $x^{adv}\!\leftarrow\!\mathcal A(x,f,K(x))$
\ENDFOR
\RETURN $\mathcal X^{adv}$
\end{algorithmic}
\end{algorithm}
\vspace{-5pt}

The framework follows the three steps in Algorithm~\ref{alg:framework}. Given an input image, Quad-CLEVER first estimates its pre-optimization attack difficulty. The estimated difficulty is then converted into an image-level optimization budget through the budget allocation policy $\pi(\cdot)$ under the total budget $B$. Finally, the allocated budget is used by the base attack to generate the adversarial example. Compared with a fixed optimization budget, the proposed framework allocates more computation to difficult inputs while avoiding unnecessary optimization on easier ones. We consider two implementations of the budget allocation policy $\pi(\cdot)$.

\subsubsection{Discrete Budget Routing}

The first implementation partitions inputs into three difficulty levels and assigns a predefined optimization budget to each level,

\begin{equation}
K(x)=
\begin{cases}
K_{\mathrm{easy}}, &
\widehat D(x;f)\le\gamma_1,\\
K_{\mathrm{med}}, &
\gamma_1<\widehat D(x;f)\le\gamma_2,\\
K_{\mathrm{hard}}, &
\widehat D(x;f)>\gamma_2,
\end{cases}
\label{eq:3bucket}
\end{equation}

where $\gamma_1$ and $\gamma_2$ are calibration thresholds estimated from a calibration set.

\subsubsection{Continuous Budget Routing}

Instead of assigning one of three predefined budgets, the second implementation directly maps the estimated difficulty to the optimization budget through a continuous monotonic function,

\begin{equation}
K(x)=
10\cdot
\mathrm{round}
\left(
\frac{
K_{\min}
+
(K_{\max}-K_{\min})
\rho(\widehat D(x;f))^\gamma
}{10}
\right),
\label{eq:continuous_routing}
\end{equation}

where $\rho(\cdot)$ is obtained through logarithmic normalization followed by percentile clipping,

\begin{align}
z(x)&=\log(\widehat D(x;f)+\eta),\\
\rho(\widehat D)&=
\mathrm{clip}
\left(
\frac{z-P_5^z}{P_{95}^z-P_5^z},
0,1
\right).
\end{align}

The logarithmic transformation mitigates the heavy-tailed distribution of difficulty scores, while percentile clipping improves robustness by limiting the influence of extreme values.
\section{Experiments}

Our experiments are designed to answer the following two research questions.

\textbf{RQ1:} \emph{Can attack difficulty be reliably estimated before adversarial optimization begins?}
To answer this question, we evaluate whether Quad-CLEVER consistently predicts the relative optimization difficulty across input samples using multiple attack algorithms and optimization cost metrics.

\textbf{RQ2:} \emph{Can pre-optimization difficulty estimation improve the efficiency of adversarial attacks?}
To answer this question, we incorporate the estimated difficulty into our difficulty-aware attack framework and evaluate whether adaptive budget allocation improves attack efficiency under a fixed computational budget.

\subsection{Experiment Setup}
\paragraph{Threat model.} We consider a white-box person-vanishing attack setting. The attacker has full access to the target detector including its architecture, parameters, gradients, and post-processing procedure. Given a clean image containing one or more correctly detected person instances, our goal is to generate an adversarial image that causes the detector to miss the target persons while satisfying a predefined $\ell_2$ perturbation constraint. Specifically, we aim to suppress the person predictions associated with the original ground-truth instances after the complete detection pipeline, including confidence filtering and non-maximum suppression. A target person instance is regarded as successfully attacked if no remaining person prediction can be matched to it under the match IoU threshold 0.5. This criterion accounts for detection regeneration, where suppressing one prediction may cause another highly overlapping bounding box to survive after non-maximum suppression. At the image level, an attack is considered successful if at least one target person instance is removed.

\paragraph{Dataset.}
Our primary datasets are BDD100K \cite{yu2020bdd100k} and EventPed \cite{zhang2024when}. BDD100K is a large-scale autonomous-driving dataset with diverse road, weather, and illumination conditions. Meanwhile, EventPed serves as a more challenging pedestrian-detection benchmark, containing approximately 9K RGB-event image pairs collected in varied outdoor scenes, including daytime, nighttime, and occluded conditions. For both datasets, we retain images containing at least one annotated person instance.

{\noindent\bf Models.}
YOLOv8s \cite{yolov8_ultralytics} is selected  as the target object detector. Specifically, we initialize the model using the official checkpoint pretrained on the COCO dataset and fine-tune it on two datasets separately. All adversarial attacks and evaluations are subsequently conducted on the fine-tuned YOLOv8s model.

{\noindent\bf Attacks.}
We evaluate the effectiveness of our difficulty-aware routing strategy on two gradient-based baseline attacks. PGD: an $\ell_p$-bounded projected gradient attack adapted to the object-vanishing objective; TOG: a targeted objectness-gradient vanishing attack following \cite{chow2020adversarial}. Both attacks are conducted under the same configuration, using an $\ell_2$ perturbation constraint with a maximum perturbation budget of $\epsilon=2.0$ and a step size $\alpha=0.005$. We apply our routing strategy to both baselines to evaluate whether it can effectively allocate attack iterations according to the estimated difficulty of each input image.

{\noindent\bf Metrics.}
The evaluation metrics are Attack Success Rate ({\bf ASR}) and the average number of attack iterations ({\bf Avg.K}). ASR measures the proportion of evaluated images for which at least one target person instance is successfully removed after confidence filtering and non-maximum suppression. Avg.K denotes the mean number of attack iterations allocated to each image and is used to quantify the computational cost of the attack. Together, these two metrics characterize the trade-off between attack effectiveness and attack efficiency.

{\noindent\bf Baseline.}
The baselines include fixed-K, which assigns the same iteration budget to all images, and bucket3 routing~\eqref{eq:3bucket}, which divides images into easy, medium, and hard groups according to Quad-CLEVER. The main evaluation focuses on our continuous log-normalized routing strategy~\eqref{eq:continuous_routing} with default $\gamma=1.0$.

{\noindent\bf Hardware.}
All experiments were conducted on a workstation equipped with RTX 3090 GPU under Ubuntu 22.04, using Pytorch 2.0 and CUDA 12.2.
\begin{table}[t]
    \centering
    \resizebox{\columnwidth}{!}{
    \begin{tabular}{llcccc}
        \toprule
        Dataset & Attack &
        \multicolumn{3}{c}{ASR gain of Log continuous (pp)} &
        $\Delta K_{\mathrm{save}}$ \\
        \cmidrule(lr){3-5}
        & & vs. Random & vs. Uniform &
        vs. Fixed $K$ &
        at same ASR \\
        \midrule
        BDD100K
        & PGD
        & \textbf{+5.78}
        & \textbf{+3.45}
        & \textbf{+3.08}
        & \textbf{+11.42} \\

        BDD100K
        & TOG
        & \textbf{+5.00}
        & \textbf{+2.52}
        & \textbf{+1.50}
        & \textbf{+2.36} \\

        EventPed
        & PGD
        & \textbf{+1.69}
        & $-0.41$
        & \textbf{+0.58}
        & \textbf{+2.25} \\

        EventPed
        & TOG
        & \textbf{+0.07}
        & \textbf{+2.67}
        & \textbf{+0.92}
        & \textbf{+2.13} \\
        \bottomrule
    \end{tabular}}
    \caption{
    ASR gains and iteration efficiency of Log continuous routing. ASR gains are measured in percentage points. Positive $\Delta K_{\mathrm{save}}$ indicates fewer iterations required to achieve the same ASR.
    }
    \label{tab:routing_gain}
    \vspace{-10pt}
\end{table}

\begin{figure}[t]
    \centering
    \includegraphics[width=\columnwidth]{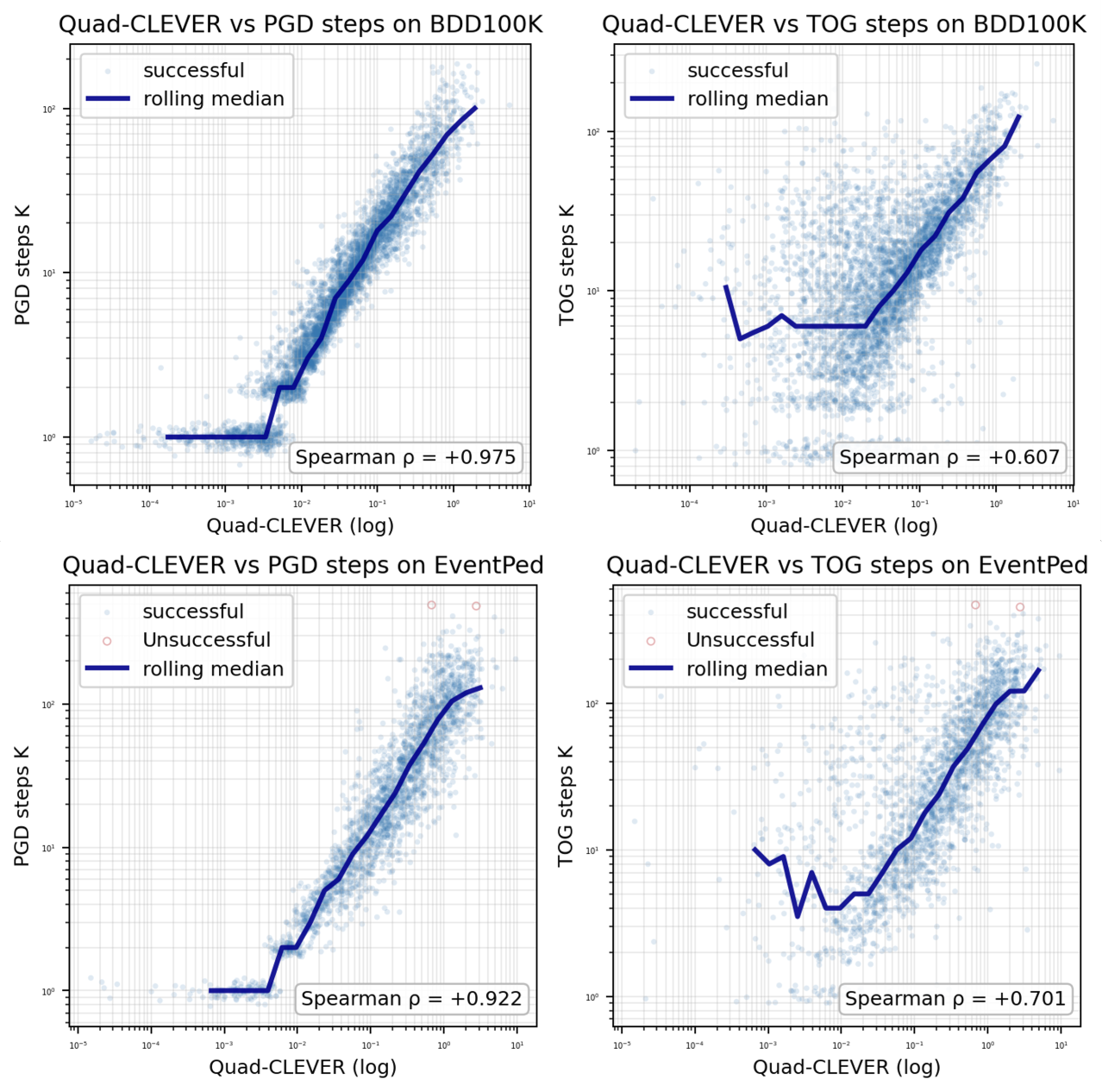}
    \caption{
    Quad-CLEVER versus the first successful attack iteration for PGD and TOG on BDD100K and EventPed. Filled points denote successfully attacked images, while hollow red points indicate images that remain unsuccessful within the maximum iteration budget. The solid curve shows the rolling median over successful images.
    }
    \label{fig:quad_attack_steps}
    \vspace{-15pt}
\end{figure}


\subsection{RQ1: Can Attack Difficulty Be Estimated Before Optimization?}

We first investigate whether Quad-CLEVER can estimate attack difficulty prior to adversarial optimization. For each image $x$, we compute the image-level Quad-CLEVER score defined in 
{Eq.~\eqref{eq:difficulty}}
, where $\mathcal{O}(x)$ denotes the set of detected person instances. We then execute each attack with a maximum budget of 500 optimization iterations and record the iteration at which the first successful attack occurs. This first successful iteration serves as an empirical measure of attack difficulty, with larger values indicating greater optimization effort. Since both Quad-CLEVER scores and attack iterations span multiple orders of magnitude, we visualize them on logarithmic axes and quantify their relationship using Spearman's rank correlation. A rolling median over successful attacks is additionally plotted to highlight the overall trend.

As shown in Fig.~\ref{fig:quad_attack_steps}, Quad-CLEVER exhibits a strong monotonic relationship with the optimization effort required by PGD. The Spearman correlation reaches $\rho=0.975$ on BDD100K and $\rho=0.922$ on EventPed. Images with small Quad-CLEVER scores are typically attacked within only a few iterations, producing a visible floor near $K=1$. As the score increases, the rolling median rises almost monotonically, eventually exceeding one hundred iterations for the most difficult samples. The consistent behavior across both datasets suggests that the observed relationship is not specific to a particular data distribution.

Quad-CLEVER also remains predictive for TOG, although the relationship is weaker than for PGD. The corresponding Spearman correlations are $\rho=0.607$ on BDD100K and $\rho=0.701$ on EventPed. Greater variability is observed among low-difficulty samples, resulting in a flatter rolling median in this region. Nevertheless, the median attack cost increases steadily with the Quad-CLEVER score, indicating that more difficult samples consistently require more optimization iterations. The weaker correlation is expected because TOG optimizes an objectness-oriented objective whose optimization trajectory is additionally affected by interactions among multiple detector hypotheses, confidence filtering, and NMS regeneration.

Overall, these results provide strong empirical evidence that Quad-CLEVER captures an image-dependent component of attack difficulty before adversarial optimization begins. The estimated score consistently predicts the relative optimization effort required by two substantially different attack algorithms across two datasets, supporting its use as a pre-optimization difficulty estimator. This property motivates its use in the subsequent difficulty-aware routing framework, where attack budgets are allocated according to the estimated difficulty.

\subsection{RQ2: Can Pre-optimization Difficulty Improve Attack Efficiency?}

The preceding experiments show that Quad-CLEVER characterizes the local robustness of person instances. We further investigate whether it can provide a difficulty ordering for allocating attack iterations. Two strategies are considered, {\bf Bucket3} from Eq.\eqref{eq:3bucket} and {\bf Log continuous} from Eq.~\eqref{eq:continuous_routing}. We additionally construct two matched baselines. {\bf Log uniform matched} applies a fixed budget close to the average budget of continuous routing, whereas {\bf Log random matched} randomly permutes the budgets assigned by continuous routing across images. The latter preserves the nominal budget distribution while removing the correspondence between the Quad-CLEVER ranking and attack strength. Under early stopping, the average number of actually executed iterations may differ slightly between the continuous and randomly permuted variants.

\begin{figure}[t]
    \centering
    \includegraphics[width=1\columnwidth]{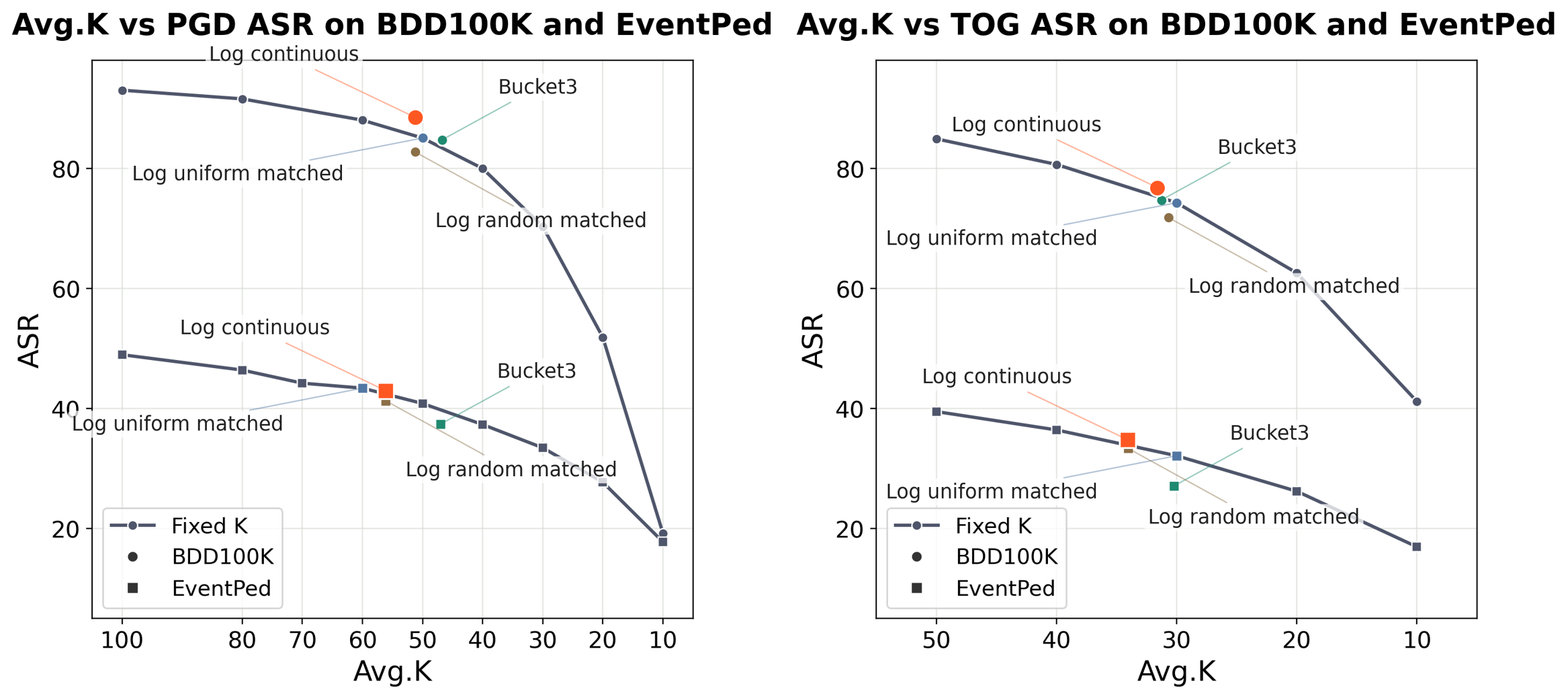}
    \caption{
    Average executed iterations versus image-level ANY ASR for PGD and TOG on BDD100K and EventPed. Fixed-$K$ results form the reference curves, while Bucket3 and Log continuous routing allocate different iteration budgets according to the Quad-CLEVER score.
    }
    \label{fig:avgk_asr}
    \vspace{-10pt}
\end{figure}

\begin{figure}[t]
    \centering
    \includegraphics[width=1\columnwidth]{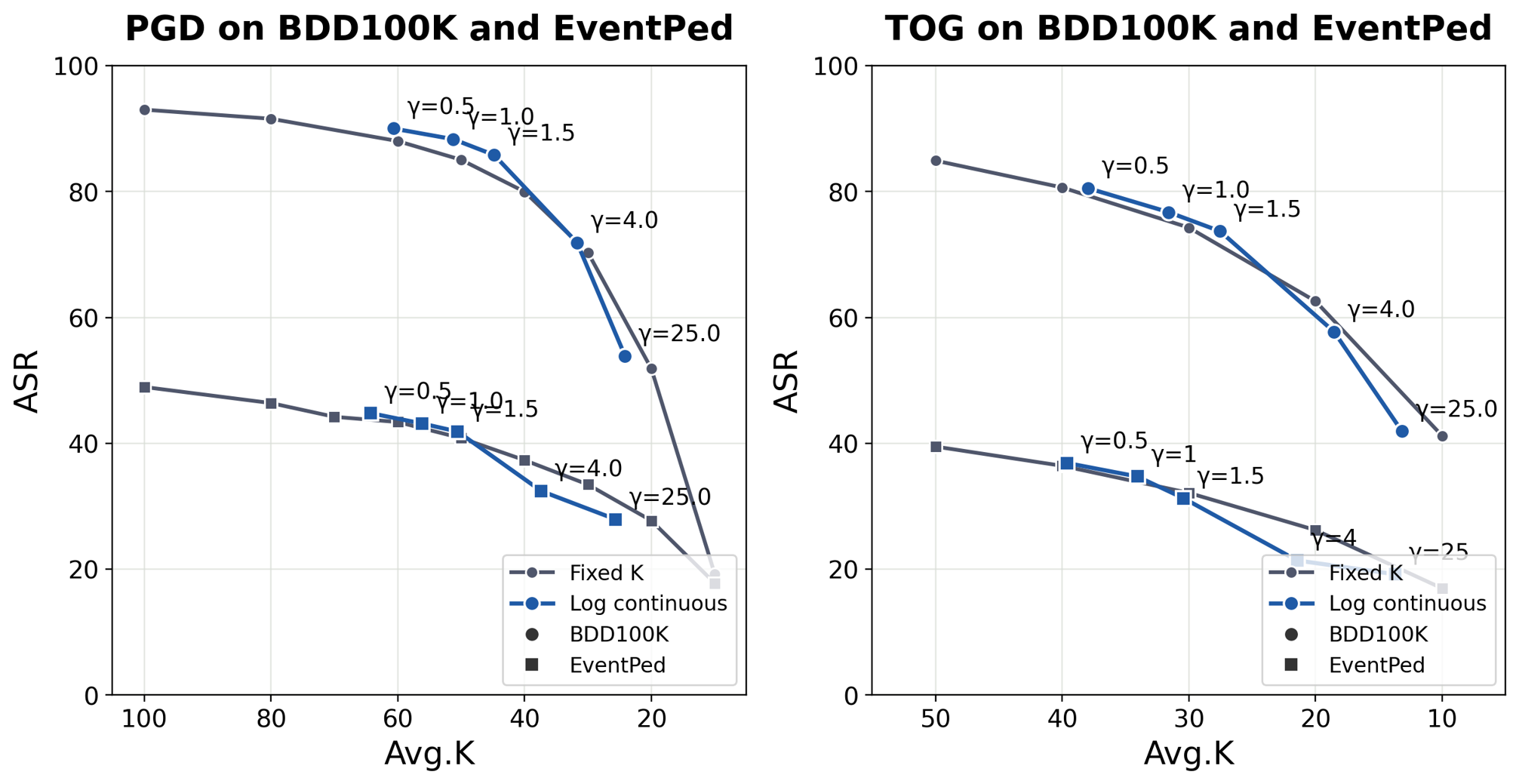}
    \caption{
    ASR–Budget Trade-off of Quad-CLEVER-Guided Routing. We vary $\gamma$ to produce continuous-routing results with different Avg.$K$ and compare them with fixed-$K$ attacks under similar average budgets. 
    }
    \label{fig:gamma_ablation}
    \vspace{-10pt}
\end{figure}

Figure~\ref{fig:avgk_asr} and Table~\ref{tab:routing_gain} demonstrate that difficulty-aware routing is stable on BDD100K. For both PGD and TOG, Log continuous routing consistently improves over random matching, uniform allocation, and fixed-$K$ attacks under the same average computational budget. The substantial gains over random matching show that the improvement is produced by assigning larger budgets to more difficult images according to the Quad-CLEVER ordering, rather than by merely changing the overall budget distribution.

The routing strategy also provides a clear efficiency improvement. At an equivalent ASR, Log continuous routing saves approximately 11.42 iterations per image for PGD and 2.36 iterations for TOG on BDD100K. Thus, the proposed routing not only increases ASR under the same average budget, but also requires fewer iterations to attain the same attack success rate. This consistent improvement across two different attacks supports the effectiveness of Quad-CLEVER as a difficulty-aware routing signal.

The gains on EventPed are smaller because the dataset is more difficult to attack, as reflected by its lower and flatter fixed-$K$ curves. Nevertheless, Log continuous routing remains above the interpolated fixed-$K$ baseline for both attacks. At the same ASR, it saves 2.25 iterations for PGD and 2.13 iterations for TOG. For PGD, routing also improves over random matching, although it remains close to the uniform baseline. For TOG, it improves over the uniform baseline and is approximately comparable to random matching. These results indicate that the difficult nature of EventPed compresses the absolute routing gains, but does not eliminate the improvement in the computation--ASR trade-off.

Overall, Quad-CLEVER enables Log continuous routing to achieve both higher ASR at a matched computational budget and lower computation at a matched ASR. The improvements are strongest and most consistent on BDD100K, while the more challenging EventPed dataset still exhibits positive efficiency gains for both PGD and TOG.

    \begin{table}[t]
    \centering
        \resizebox{1\columnwidth}{!}{
        \begin{tabular}{lccccccccc}
        \toprule
        & \multicolumn{3}{c}{Top-20\% Overlap (\%)}
        & \multicolumn{6}{c}{Top-10\% $\rightarrow$ Top-20\% (\%)} \\
        \cmidrule(lr){2-4}
        \cmidrule(lr){5-10}
        Dataset
        & Q--P & Q--T & P--T
        & Q$\rightarrow$P & P$\rightarrow$Q
        & Q$\rightarrow$T & T$\rightarrow$Q
        & P$\rightarrow$T & T$\rightarrow$P \\
        \midrule
        BDD100K
        & 85.17 & 66.09 & 70.57
        & 98.39 & 98.62 & 87.82 & 76.55 & 94.02 & 79.31 \\
        EventPed
        & 73.13 & 64.79 & 80.21
        & 82.50 & 82.92 & 75.42 & 69.17 & 95.42 & 85.83 \\
        \bottomrule
        \end{tabular}}
        \caption{Consistency between difficulty rankings produced by Quad-CLEVER (Q), PGD (P), and TOG (T). $A_{10}\rightarrow B_{20}$ denotes the proportion of $A$'s top-10\% samples contained in $B$'s top-20\%.}
        \label{tab:consistency}
        \vspace{-10pt}
    \end{table}

\subsection{Ablation study}
{\bf Effectiveness of Quad-CLEVER-guided budget allocation.} We vary $\gamma$ only to obtain continuous-routing results at different average iteration budgets, allowing a direct comparison with fixed-$K$ attacks at similar Avg.\ $K$. As shown in Fig.~\ref{fig:gamma_ablation}, Quad-CLEVER-guided routing generally achieves a higher attack success rate than uniformly assigning the same number of iterations to every image. The improvement is most evident when the routing preserves sufficient budget for difficult samples: easy images receive fewer iterations, while the saved computation is redirected to harder ones. When $\gamma$ becomes excessively large, most budgets are compressed toward $K_{\min}$, weakening this adaptive allocation and causing the advantage over fixed-$K$ to diminish or disappear. These results verify that Quad-CLEVER provides a meaningful difficulty ordering and can improve the allocation of attack computation across images.

{\noindent\bf Consistency across attack algorithms.}
We investigate whether Quad-CLEVER is confined to a specific attack algorithm. We rank images using Quad-CLEVER and the optimization steps required by standalone PGD and TOG. We compare both the overlap between their top-20\% difficult subsets and whether the most difficult top-10\% samples identified by one method fall within the top-20\% of another. As shown in Table~\ref{tab:consistency}, Quad-CLEVER is highly consistent with the empirical difficulty measured by PGD and also maintains clear agreement with TOG. All difficult-subset overlaps are substantially above the random reference, while the strong PGD--TOG agreement further indicates that many difficult images are shared across attack algorithms. These results suggest that Quad-CLEVER captures an attack-agnostic property of image difficulty.

    \begin{table}[t]
        \centering
        \resizebox{\columnwidth}{!}{
        \begin{tabular}{llcccc}
        \toprule
        Dataset & Attack & All & Person=1 & Person=2 & Person=3 \\
        \midrule
        BDD100K & PGD & 0.9754 & 0.9789 & 0.9770 & 0.9825 \\
        BDD100K & TOG & 0.6063 & 0.7437 & 0.5996 & 0.5019 \\
        EventPed & PGD & 0.9289 & 0.8793 & 0.9240 & 0.9360 \\
        EventPed & TOG & 0.7349 & 0.6818 & 0.7148 & 0.7487 \\
        \bottomrule
        \end{tabular}
        }
        \caption{Spearman correlation between Quad-CLEVER and attack consumption for different numbers of persons.}
        \label{tab:person_num}
        \vspace{-10pt}
    \end{table}

{\noindent\bf Effect of the Number of Persons.} 
To examine whether the effectiveness of Quad-CLEVER is affected by the number of target persons, we divide the images according to their clean person count and separately calculate the Spearman correlation between Quad-CLEVER and the required attack iterations in Table~\ref{tab:person_num}. For PGD, Quad-CLEVER maintains a consistently strong correlation across different person counts on both datasets, indicating that its difficulty ordering is not dominated by the number of targets. The correlations with TOG remain positive but show greater dataset-dependent variation. In particular, the correlation decreases as the person count increases on BDD100K, whereas it remains stable or slightly improves on EventPed. This difference may arise because the image-level minimum Quad-CLEVER score characterizes the most vulnerable target, while the optimization cost of TOG can additionally depend on interactions among multiple persons. Overall, these results demonstrate that Quad-CLEVER captures attack difficulty beyond the simple effect of person count.



\section{Conclusion}

In this work, we investigated whether person-vanishing attack difficulty contains an input-dependent structure that can be characterized before adversarial optimization. To provide observable evidence, we proposed Quad-CLEVER, an efficient geometry-based estimator derived from a directional second-order approximation of the detector margin. Experiments show that Quad-CLEVER consistently predicts the relative optimization effort required by different attack algorithms, indicating that attack difficulty is not solely an outcome of a particular optimization process. Based on this ordering, we further developed difficulty-aware routing strategies that adaptively allocate fewer iterations to easier images and more computation to harder ones. The resulting routing strategies improve the trade-off between attack success and optimization cost across multiple attacks and datasets. Overall, our findings demonstrate that person-vanishing attack difficulty can be meaningfully estimated from clean inputs and that exploiting this information enables more efficient adversarial optimization for object detectors.

\bibliography{aaai2027}



\end{document}